*Research Article*

# Multiclass Classification Procedure for Detecting Attacks on MQTT-IoT Protocol

**Hector Alaiz-Moreton** ,[1] **Jose Aveleira-Mata** [ID],[2] **Jorge Ondicol-Garcia,**[2]
**Angel Luis Muñoz-Castañeda,**[2] **Isaías García,**[1] and **Carmen Benavides**[1]

[1]*Escuela de Ingenierías, Universidad de León, 24071 León, Spain*
[2]*Research Institute of Applied Sciences in Cybersecurity (RIASC) MIC, Universidad de León, 24071 León, Spain*

Correspondence should be addressed to Hector Alaiz-Moreton; hector.moreton@unileon.es

The large number of sensors and actuators that make up the Internet of Things obliges these systems to use diverse technologies and protocols. This means that IoT networks are more heterogeneous than traditional networks. This gives rise to new challenges in cybersecurity to protect these systems and devices which are characterized by being connected continuously to the Internet. Intrusion detection systems (IDS) are used to protect IoT systems from the various anomalies and attacks at the network level. Intrusion Detection Systems (IDS) can be improved through machine learning techniques. Our work focuses on creating classification models that can feed an IDS using a dataset containing frames under attacks of an IoT system that uses the MQTT protocol. We have addressed two types of method for classifying the attacks, ensemble methods and deep learning models, more specifically recurrent networks with very satisfactory results.

## 1. Introduction

The "Internet of Things" (IoT) describes many different systems and devices that are constantly connected to Internet, giving information from their sensors or interacting with their actuators. By 2020 it is estimated that there will be 4.5 billion IoTs joining the Internet [1]. These devices have special features, such as a low computing capacity and the use specific lighter protocols. This makes IoT devices more efficient, smaller, and less energy consuming; however these low settings reduce their encryption capacity. These heterogeneous systems and networks offer new challenges in cybersecurity, such as new vulnerabilities and anomalies [2, 3]. One of the most important attacks in recent years, the Mirai botnet, exploited these vulnerabilities by carrying out distributed denial of service attacks infecting IoT devices and attacking with as many as 400,000 simultaneously connected devices [4].

One way of improving network security is the use of Intrusion Detection Systems (IDS). IDS are one of the most productive techniques for detecting attacks within a network.

This tool can detect network intrusions and network misuses by matching patterns of known attacks against ongoing network activity [5]. With this purpose, our focus is to develop an IDS with machine learning models for the IoT. IDS use two different detection methods: signature-based detection and anomaly-based detection. Signature-based detection methods are effective in detecting well-known attacks by inspecting network traffic for specific patterns. Anomaly-based detection systems identify attacks by monitoring the behaviour of the entire system, objects, or traffic and comparing them with a predefined normal status [6].

Machine learning techniques are used to improve detection methods, by creating new rules automatically for signature-based IDS or adapting the detection patterns of anomaly-based IDS. These anomaly-based IDS have had good results in qualifying frames that may be under attack [7], and they are effective even in detecting zero-day attacks [8].

To build a machine learning classifier it is necessary to use a dataset. Within the network intrusion detection there are some well-known datasets that are used to feed IDS with machine learning techniques [9]. As there are no public datasets based on network traffic using IoT protocols, we have used a dataset that has been created in our previous research (the dataset is available in https://joseaveleira.es/dataset. ©® reg#LE-229-18). The main focus of this paper is on the three different machine learning techniques that classify three different attacks and normal frames at the same time using our IoT environment dataset.

## 2. Related Work

There are several approaches for the detection of anomalies in traditional networks using machine learning. The most widely used datasets are the KDD99 [10] and NSL-KDD Dataset [11] (an improved version of KDD'99). These datasets contain traffic captured on the TCP protocol and collect different types of attacks. Based on these datasets, some models have been developed for anomaly detection using a Support Vector Machine and Random Forest [12, 13]. Another technique used on this dataset is K-Centroid clustering, whose objective is to improve the performance of other models [14]. There are also ways of upgrading these datasets, such as balancing classes to increase the models' prediction accuracy, which improves their performance [15].

Other detection techniques with good results are the use of Fuzziness based semisupervised learning getting an accuracy of 84 [16] and also obtaining good results analyzing the network traffic using sequential extreme learning machine with accuracies around 95 [17]. These good results indicate that machine learning is a good approach to improve the detection of intrusions in the network layer.

The machine learning methods are based on deep learning [18]. There are many approaches for solving anomaly detection using deep learning. One proposed way is to use the Deep Belief Network (DBN) as a feature selector on the KDD dataset, combined with a SVM that classifies the attacks [19, 20]. Another proposed method is to use deep learning models as feature selectors using the Fisher Score, a classical statistic method, combined with an autoencoder to reduce the dimensions of the data and extract the highest-valued features [21]. Deep learning models are used as classifiers too. Understanding that the temporary data sequence of network attacks is important, the Long Short Term Memory (LSTM) network, a variant of recurrent networks, has been used to classify the KDD's attacks [22].

As regards the IoT IDS, there is an approach that uses fog computing combined with mobile edge computing. Using this combination, a numerical simulation is made for the NSL-KDD dataset, where it has been demonstrated that this type of IDS has a good performance both in accuracy and time dependence [23]. Using this dataset, there is also an IDS based on rules which rules are modified using machine learning KNN and SVM techniques [24]. Because the research into IDS schemes for IoT is still incipient, the proposed solutions do not cover a wide range of attacks or IoT technologies [25].

There are other more recent datasets such as the AWID [26] which collects TCP frames of data from a WLAN network over which several attacks were made on 802.11 security mechanism through which a study on Wi-Fi intrusions was made using a neural network classifier [27]; another current dataset is the CICIDS2017 [28] used to validate the detection algorithms on which training has been carried out with recurrent neural networks [29].

This research is based on a dataset specialized in a protocol implemented in IoT environments to detect specific vulnerabilities. It is specialized in an IoT protocol where does not exist dissection of traffic ready to use with research purposes.

## 3. Methods and Materials

This section describes the methods, from a theoretical point of view as well as the materials used for implementing the experiments.

*3.1. MQTT Dataset.* In order to classify anomalies in an IoT environment, we built a dataset using MQTT, which is a publish-subscribe-based messaging protocol. It is a light protocol widely used in IoT [30].

The MQTT's architecture follows a star topology, with a central node that functions as a server or broker. The broker is in charge of managing the network and transmitting. The communication is based on topics created by the client that publishes the message and the nodes that wish to receive it must subscribe to it. The communication can be one to one, or one to many.

This dataset has been obtained in a test environment with several sensors, actuators, and a server. This server hosts the management application, also working as the broker that manages the messages of the MQTT protocol. The scheme of the environment is detailed in the Figure 1.

We carried out several attacks against the MQTT protocol in the test environment. We captured these attacks at the network level along with all generated traffic. The attacks carried out were as follows:

(i) DoS: denial of service is one of the most common attacks on the Internet [31]. In the case of the MQTT protocol, the broker is attacked by saturating it with a large number of messages per second and new connections. Using the MQTT-malaria program [32], this program is used for testing the scalability and load testing utilities for MQTT environments.

(ii) Man in the middle (MitM) consists of intercepting the messages between two communication points in an attempt to modify the content; in this case it is done between a sensor and the broker by modifying the sensor data. To carry out the attack we used the distribution Kali Linux and the tool Ettercap.

(iii) Intrusion: taking into account the characteristics of the MQTT protocol, this attack consists of using the well-known port (1883) for this protocol and a command that uses the special character "#" can be used by an external attacker for knowing the active topics available for being subscripted. To find out which topics a client outside the system [33].

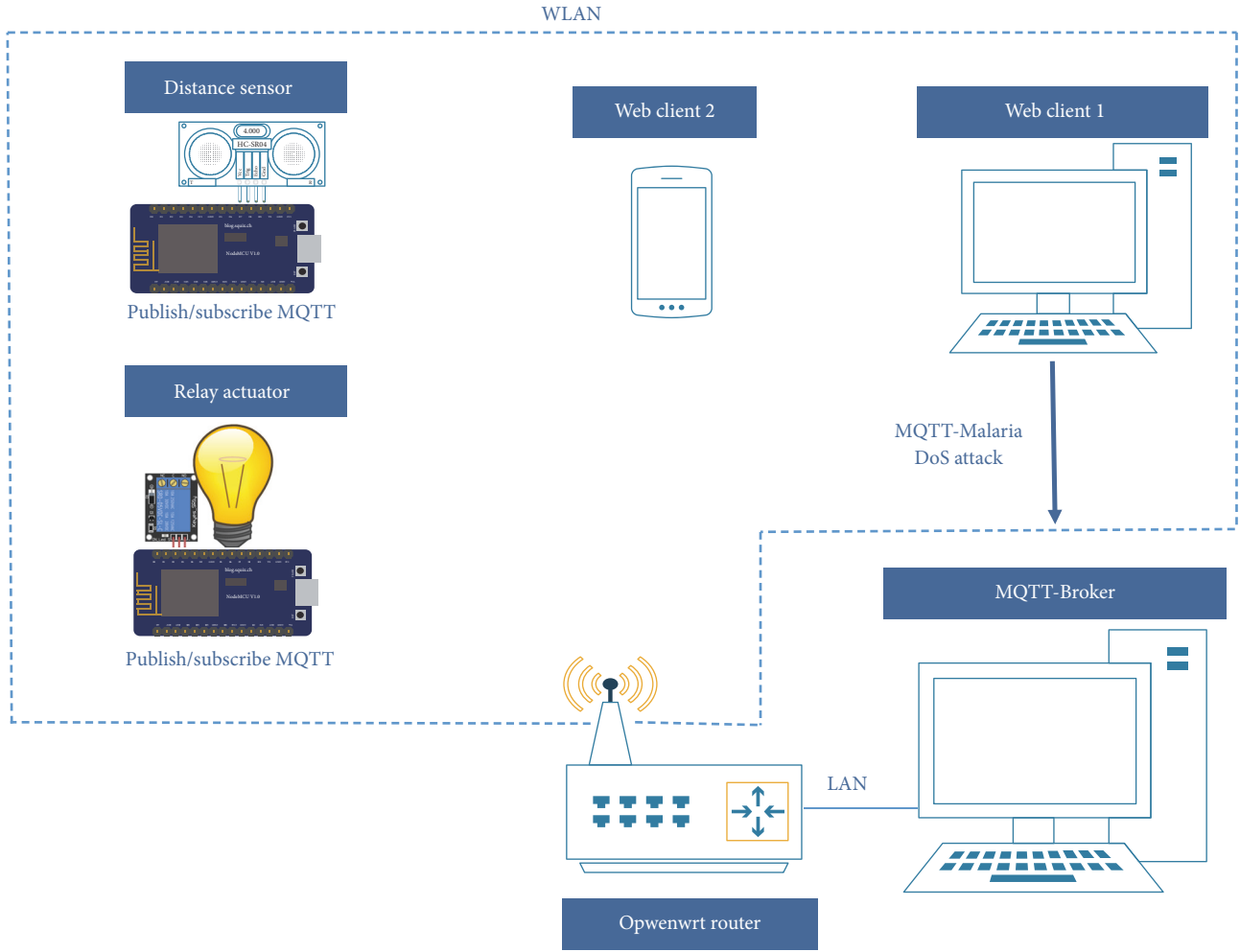

FIGURE 1: Test environment schema.

The relevant fields and the protocol are selected after capturing the network traffic in the system. All communication frames are tagged to show whether one of them is under attack or is normal. There are three CSV files generated, one for each attack, all of them being part of the dataset used. Selecting features and labeling the frames indicating whether or not they are under attack enable supervised learning techniques to be used on this dataset.

The features of the data set are as follows:

(i) DoS.csv that contains the capture of 94.625 frames and of which 45.513 are under attack traffic and 49.112 are normal traffic.

(ii) MitM.csv that contains 110668 frames with 3855 under man in the middle attack and 106.813 normal traffic frames.

(iii) Intrusion.csv with 80,893 total frames with 1898 under attack and 78,995 normal traffic frames.

*3.2. Classification Methods.* We have chosen XGBoost because other research delivered good results like [34–36]. We have also chosen recurrent networks for our experiments because of the importance of time in network attacks [22], as frames are produced sequentially, and the sequence and time between frames provide relevant information for detecting an attack. We shall go into our classification models in more detail in the following sections.

*3.2.1. XGBoost (Gradient Boosting).* Gradient boosting systems build additive models in a forward way through steps, allowing the optimization of arbitrary differentiable loss functions. In each forward step, regression trees are fitted onto the negative gradient of the binomial or multinomial loss function [37]. XGBoost stands for Extreme Gradient Boosting [38]. It is a scalable machine learning system for tree boosting which optimizes many systems and algorithms, such as a tree learning algorithm that handles sparse data, handling instance weights in approximate tree learning or exploiting out-of-core computation. For the implementation of this method, we are using the XGBoost library for Python [39].

*3.2.2. Recurrent Neural Networks.* Recurrent neural networks are a variant of neural networks designed for highly sequenced problems. RNN contain cycles that feed the network activation from a previous time step as inputs into the network, influencing predictions at the current time step. The addition of cycles gives the RNN a new dimension, where instead of mapping only inputs to outputs, the network will learn a mapping function for the inputs to an output over time. One of the main disadvantages of this kind of network is the training problems, such as a vanishing gradient and exploding gradients [40]. These problems can be addressed through variations in the neurons, such as GRU or LSTM cells.

*LSTM Recurrent Network.* LSTM networks have a unique formulation that allows them to prevent the problems of scaling and training of the vanilla RNN, avoiding the back propagation error that either blows up or decays exponentially. An LSTM layer consists of a set of recurrently connected blocks, known as memory blocks that are the computational units of the LSTM network [41]. These cells are made up of weights and gates. Each memory block contains one or more recurrently connected memory cells and three multiplicative units: input, output, and forget gates. The gates allow the information flows to interact with the cells. The forget gate and the input gate update the internal state, while the output gate is the final limiter of the cells' output. These gates and the consistent data flow called CEC, or constant error carrousel, keep each cell stable [42]. For the implementation of this network, we are using the Keras framework for Python with the Tensorflow backend, using GPU processing and an improved-performance version that uses Cudnn (CudnnLSTM) [43].

*GRU Recurrent Network.* Just like LSTM networks, GRU networks have a structure and formulation that improve the vanilla RNN. GRU was first proposed in Cho et al. [44] as an alternative to the LSTM to capture dependencies of different time scales adaptively. The only difference between these networks is the procedure for updating the CEC. It is similar to the LSTM but with one difference, the GRU units have no mechanism for controlling the exposure to which this data flow is submitted [45]. This lack of control mechanisms makes the GRU units faster than the LSTM and more adaptable to the changes in the time flow [46]. For the implementation of this network, we are using the Keras framework for Python with the Tensorflow backend, using GPU processing and an improved-performance version that uses Cudnn (CudnnGRU) [43].

*3.3. Optimization Methods.* SGD methods are iterative methods used for optimizing an objective function. Adam is based on adaptive estimates of lower-order moments. This method is simple to implement and computationally efficient and has few memory requirements. It does not change as a result of the diagonal rescaling of the gradients and works well for problems that are large in data and/or parameters. Adam is also appropriate for changing objectives and problems with very noisy and/or sparse gradients. The hyperparameters of this method are easy and intuitive to understand and usually require little tuning [47]. Based on Adam we have Nadam, which is an Adam version applying Nesterov momentum

[48]. We have tested RMSprop, Adam, and Nadam, all of which are stochastic gradient descent methods, but we got our best results using Nadam to optimize our loss. Nadam brings more speed in learning in each minibatch step. Nadam gave us better results because we have a complex net and fewer epochs. We needed a faster loss function optimizer to learn more in fewer epochs, without fearing a fast Decay into overfitting.

*3.4. Batch Normalization.* This method works by making normalization a part of the model architecture and carrying out a normalization step for each training minibatch. It addresses the problem of the internal covariate shift, brought about by the values of the input layers' changing during training. This problem requires low learning rates and a careful parameter initialization and makes it harder to train models with saturating nonlinearities. Batch Normalization allows us to use higher learning rates and pay less attention to initialization. It can also act as a regularizer, in some cases eliminating the need for other regularization techniques [49].

*3.5. Evaluation Metrics.* Metrics evaluate the performance of a machine learning model. Every metric measures the efficiency in a different way, so we use several metrics for our models in order to obtain a more accurate view.

*3.5.1. Multiclass Logarithmic Loss and Categorical Cross Entropy.* The logarithmic loss metric measures the performance of a classification model in which the prediction input is a probability value of between 0 and 1. Its formula is as follows:

$$-\left(y * \log\left(y_{pred}\right) + (1 - y) * \log\left(1 - y_{pred}\right)\right) \quad (1)$$

where $y \in [0, 1]$ is the known label and $y_{pred} \in [0, 1]$ is the prediction of the model. Logarithmic loss and cross entropy in machine learning when calculating error rates of between 0 and 1 lead to the same thing. The cross-entropy formula is as follows:

$$H(p, q) = -\sum_{x} \left(p(x) * \log(q(x))\right) \quad (2)$$

If $p \in [y, 1 - y]$ and $q \in [y_{pred}, 1 - y_{pred}]$,

$$-\left(y * \log\left(y_{pred}\right) + (1 - y) * \log\left(1 - y_{pred}\right)\right) \quad (3)$$

The same formula is applied in both situations. We can extend the logarithmic loss to multiclass problems, given the true labels of a set of samples encoded as a 1-of-K binary indicator matrix $Y$, where $y_{i,k} = 1$ if sample $i$ has label $k$ taken from a set of $K$ labels. Let $Y_{pred}$ be a matrix of probability estimates, with $ypred_{i,k} = Pr(t_{i,k} = 1)$:

$$L_{log}(Y, Y_{pred}) = -\frac{1}{N} \sum_{i=0}^{N-1} \sum_{k=0}^{K-1} y_{i,k} * \log\left(ypred_{i,k}\right) \quad (4)$$

### 3.5.2. Multiclass Classification Error Rate.
The multiclass error rate is the percentage of misclassifications made by the model:

$$\frac{p_{wrong}}{P} \tag{5}$$

### 3.5.3. F-Beta Score.
The F-beta score is the weighted harmonic average of precision and recall, obtaining its best value at 1 and its worst value at 0. The $\beta$ parameter determines the weight of precision in the combined score. $\beta < 1$ lends more weight to precision, while $\beta > 1$ favors recall.

$$F_\beta = \left(1 + \beta^2\right) * \frac{precision * recall}{(\beta^2 * precision) + recall} \tag{6}$$

### 3.5.4. Categorical Accuracy.
The calculation of the average accuracy rate across all predictions made for a multiclass problem is made using the following formula:

$$\frac{1}{N} \sum_{i=0}^{N-1} Equals\left(\text{argmax}\left(y\right), \text{argmax}\left(p\right)\right) \tag{7}$$

### 3.6. Dropout.
Dropout is used to prevent overfitting. It works by randomly dropping units and their connections from the neural network during training. This prevents network units from adapting too much to a problem [50].

## 4. Experiments

Our experiments are based on three datasets, one for each attack. Before joining them, we balanced each one of them to reduce the huge differences between all of the classes. We balance the classes of each dataset using the resample method provided by Scikit-learn [51]. Once all of the datasets were balanced, we put them together to build a multiclass dataset. With the complete dataset ready, we chose the most representative features using a Feature Importance (FIM) report system. Our FIM algorithm is a hybrid method based on the mutual information function and it is composed by two routines; one corresponding to a filter process (based on the minimum-redundancy-maximum-relevance) and another corresponding to a wrapper process, where we used several models like SVM, Decision Trees, or Random Forests. This method confronts each feature of the dataset against the target feature. Choosing the highest values gives us the most important features for each set of data. After that, we confront each of the variables chosen in pairs between them and then we delete the highest-valued features to decrease redundancy. We also have to prepare our custom metric F-beta score. Taking into account the F-beta formula presented on the paper, we select beta = 1 to increase the value of the recall variable. The recall is the amount of data well classified in both parts, referring to the amount of present false positives and negatives. The classifications problems, in networks specifically, have many problems with false positives and negatives, so giving more value to the metric can make it more sensitive to these failures and could give us a more accurate vision. Immediately after that, we prepare the categorical values in order to make it possible to train both recurrent neural networks Box 1.

Finally, we set four timesteps for both recurrent LSTM and GRU networks, transforming the inputs into tensors made up of samples, timesteps, and features Box 2.

A hyperparameter search on recurrent networks is computationally expensive, so we have chosen their hyperparameters depending on logs of training, increasing the width and length of the network or increasing the periods if it is underfitted or by applying a batch normalization, dropout or reducing the length, width, or epochs to reduce overfitting. Now, we will describe our three different classification methods in greater detail.

### 4.1. XGBoost.
We define the XGBoost model for our problem, highlighting the four types that we wish to classify and specifying both the tree method and the booster. We use a version of the tree method called the XGBoost fast-histogram algorithm. This method is much faster and uses considerably less memory than other methods [52], but it needs a specific version of CUDA to work. We also highlighted the evaluation metrics that we wanted to use (multiclass logarithmic loss and multiclass classification error rate), the number of threads and the number of estimators in the model. We have applied a grid search with a threefold cross validation to take the best parameters of the model. These are the parameters we wish to tune in Box 3.

This grid search gives us a set of parameters that perform best on the problem in Box 4.

### 4.2. Recurrent LSTM.
For our LSTM network, we first compile some of the parameters of the net, setting the loss, the optimizer, and the metrics. Our loss is a variant

```
...
le_Msg = LabelEncoder()
dataset_combined['mqtt.msg'] = le_Msg.fit_transform
(dataset_combined['mqtt.msg'].astype(str))
...
```

Box 1

```
input_timesteps = 3
features = 11
X_train = scaler.fit_transform(X_train)
X_test = scaler.fit_transform(X_test)
#Three timesteps plus the actual one
X_train = X_train.reshape
(X_train.shape[0], input_timesteps+1, features)
X_test = X_test.reshape
(X_test.shape[0], input_timesteps+1, features)
```

Box 2

TABLE 1: Results of evaluation metrics for XGBoost.

| Model | M. logarithmic loss | M. classification error rate |
|---|---|---|
| XGBoost-Train | 0.075348 | 0.024753 |
| XGBoost-Test | 0.079451 | 0.025651 |

```
param = ['max-depth': [1, 5, 10, 20, 25],
'learning-rate': [0.4, 0.6, 0.8],
'min-child-weight': [1, 5, 10],
'gamma': [0.5, 1, 1.5, 2, 5],
'sub-sample': [0.6, 0.8, 1.0],
'col-sample-by-tree': [0.6, 0.8, 1.0]]
```

Box 3

```
'learning-rate': 0.4,
'gamma': 0.5,
'min-child-weight': 10,
'col-sample-by-tree': 1.0,
'max-depth': 5,
'sub-sample': 0.8
```

Box 4

```
model.compile(loss='categorical_crossentropy',
optimizer='Nadam',
metrics=[metrics.categorical_accuracy, fbeta])
```

Box 5

```
history = model.fit(X_train, y_train, batch_size=128,
validation_split=0.1, epochs=15,
verbose=2, callbacks=[tb_LOG])
```

Box 6

of cross entropy for multiclassification called categorical cross entropy. We are using the unmodified version of the Nadam optimizer. We tested Adam, RMSprop, and Nadam, establishing that Nadam is more efficient than Adam and RMSprop for our model. We set the metrics categorical accuracy and f-score to measure the model's accuracy and reliability, setting $\beta$ parameter for Fbeta-score metric at 2 in Box 5.

We fit the model with our data, using a batch size of 128 and 15 epochs. We use a 10% validation split to validate the results of the training in Box 6.

We use an Encoder-Decoder approach for our LSTM network. We also use a CUDA version of the LSTM cell from the Keras library [43]. In order to avoid overfitting, we used dropout, setting its value between 0.3 and 0.4 depending on the size of the previous layers, and batch normalization to control exploding gradients and speed up the training process in Box 7.

*4.3. Recurrent GRU.* For our GRU network, we compile the parameters of the net, setting the loss, the optimizer, and the metrics. We have set some of the GRU net parameters similar to our LSTM net. Our loss is a variant of cross entropy for multiclassification called categorical cross entropy. We are using the unmodified version of the Nadam optimizer. As we did on LSTM, we tested both Adam and Nadam, finding that Nadam is more efficient than Adam for our model. We set the categorical accuracy and f-score of the metrics to measure the model's accuracy and reliability, setting $\beta$ parameter for F-beta-score metric at 2 in Box 8.

We fit the model with our data, using a batch size of 256 and 17 epochs. We use a 10% validation split to validate the results of the training in Box 9.

We use a linear structure approach for our GRU network. We also used a CUDA version of GRU cell from the Keras library [43]. In order to avoid overfitting, we used dropout, setting its value between 0.2 and 0.3 depending on the size of the previous layers, and batch normalization to control exploding gradients and speed up the training process. Because of the GRU cell design, it needs more time to learn than LSTM, although it is faster. This also affects dealing with the overfitting, needing fewer dropout values for the GRU net in Box 10.

## 5. Results and Discussion

Once we had the results of the search for the XGBoost, we trained the model and tested it on our data, with the following results as detailed in Figure 2 and Table 1.

Our LSTM and GRU networks gave us the following results for training and validation (Figures 3, 4, 5, 6 and Table 2).

In our previous work, the ensemble methods gave us better accuracies than the linear methods on the three datasets separately; i.e., for DoS, the best accuracy achieved was 0.99377 using a random forest model and Boosting Gradient achieved 0.99373, while SVM achieved 0.99023, taking into account the best two models and the worst. The difference was smaller for DoS, but on intrusion and MitM the difference between these two types of method was higher. Specifically, on intrusion Random Forest and Boosting Gradient they got 0.95294 and 0.95385, respectively, while SVM got 0.93031. The results were similar for intrusion. In our experiments on this paper, XGBoost achieved the highest accuracy. This result confirms that ensemble methods achieve higher accuracies

```
...
model.add(CuDNNLSTM(128, return_sequences=True))
model.add(BatchNormalization())
model.add(Dropout(0.3))
model.add(CuDNNLSTM(128, return_sequences=True))
model.add(BatchNormalization())
model.add(CuDNNLSTM(256, return_sequences=True))
model.add(BatchNormalization())
model.add(Dropout(0.4))
model.add(CuDNNLSTM(256, return_sequences=True))
model.add(BatchNormalization())
...
```

Box 7

Table 2: Results of evaluation metrics for LSTM and GRU.

| Model | Categorical Cross Entropy | Categorical Accuracy | F-beta Score |
|---|---|---|---|
| LSTM-Train | 0.2093 | 0.9276 | 0.9148 |
| GRU-Train | 0.1334 | 0.9554618 | 0.952418 |
| LSTM-Validation | 0.1821 | 0.9337 | 0.9328 |
| GRU-Validation | 0.1280 | 0.960836 | 0.95777 |

```
model.compile(loss='categorical_crossentropy',
optimizer='Nadam',
metrics=[metrics.categorical_accuracy, fbeta])
```

Box 8

```
history = model.fit(X_train, y_train, batch_size=256,
validation_split=0.1, epochs=18,
verbose=2, callbacks=[tb_LOG])
```

Box 9

and less loss than other linear models or neural networks such as SVM, GRU, and LSTM for problems involving attacks on IoT networks.

Multiclass classification problems tend to be more complex than binary problems, making getting better results harder for these problems. We had similar results in both experiments on ensemble models when classifying, where we maintain the highest metrics and results. Focusing on our GRU and LSTM models, we had better results overall using deep learning than using linear models, but we had worse results than ensembles. LSTM got worse result than the SVM's DoS model and slightly better results than the SVM model for intrusion and MitM. GRU performed better, getting worse results than the SVM's DoS, but better than SVM for intrusion and MitM.

Even though we dealt with imbalance, there are still huge differences between classes. This may have affected the accuracy in some of our models negatively, specifically the GRU and LSTM. We have been able to maintain a good result by taking the sequencing of the problem into account.

## 6. Conclusion

IoT systems have been growing in recent years and are expected to increase considerably. The special features of these devices make the network technologies more heterogeneous than traditional networks, presenting new challenges to cybersecurity. Taking into account the fact that IDS are an important security barrier that can detect intrusions and security risks in the network quickly, we propose models for the detection of attacks in IoT environments that can provide an IDS oriented for IoT. We use specific datasets with particular attacks for these systems, specifically the MQTT protocol. In this case, machine learning techniques can be used to classify the frames that an IDS can assign as attack or normal. We chose the LSTM, GRU, and XGBoost models for our classification problem. We selected these recurrent models because of the importance of time and sequencing in network attacks. We picked XGBoost because the structure of the problem benefits the hierarchical ensemble method's performance, enabling them to achieve the highest accuracies. All these three classification methods are very efficient, with GPU implementations. Ensemble methods obtained the highest results, and deep learning models achieved better results in general than linear models, but not as good as ensemble methods. These models can be used for future work in which an IDS is fed with a model. This IDS will be implemented in a standard computer

```
...
model.add(CuDNNGRU(128, return_sequences=True))
model.add(BatchNormalization())
model.add(Dropout(0.2))
model.add(CuDNNGRU(256, return_sequences=True))
model.add(BatchNormalization())
model.add(CuDNNGRU(256, return_sequences=True))
model.add(BatchNormalization())
model.add(Dropout(0.3))
model.add(CuDNNGRU(256, return_sequences=True))
model.add(BatchNormalization())
...
```

Box 10

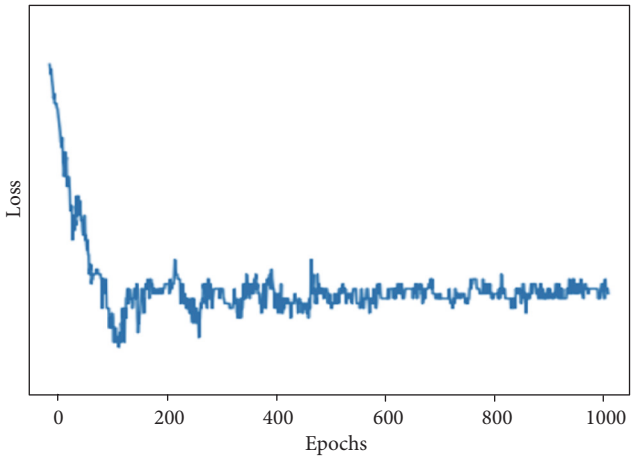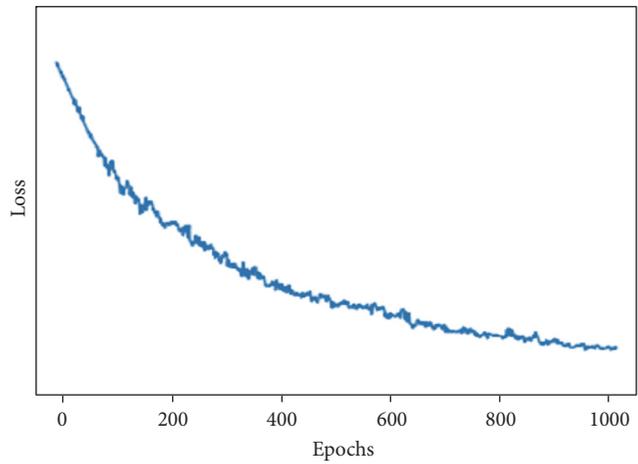

FIGURE 2: Training and testing graphics for XGBoost.

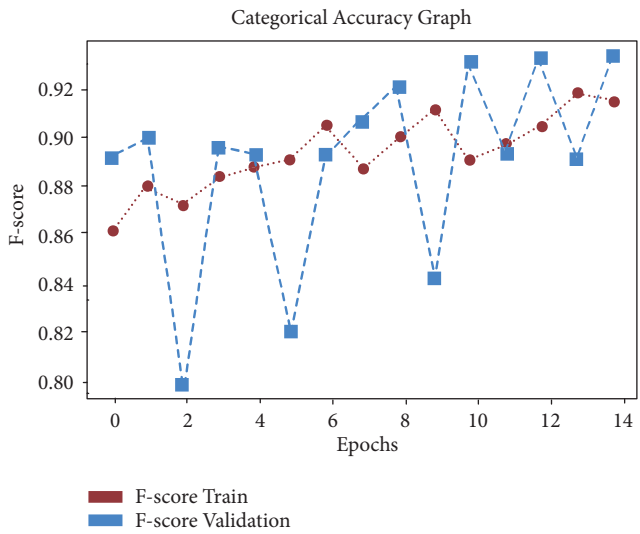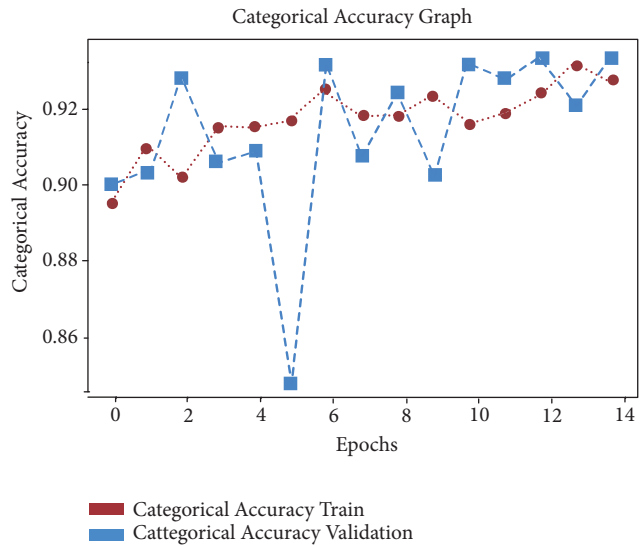

FIGURE 3: F-beta score and categorical accuracy LSTM.

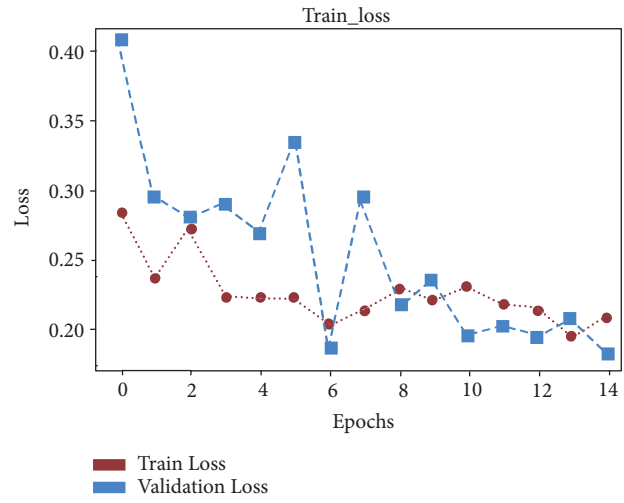

Figure 4: Categorical cross-entropy LSTM.

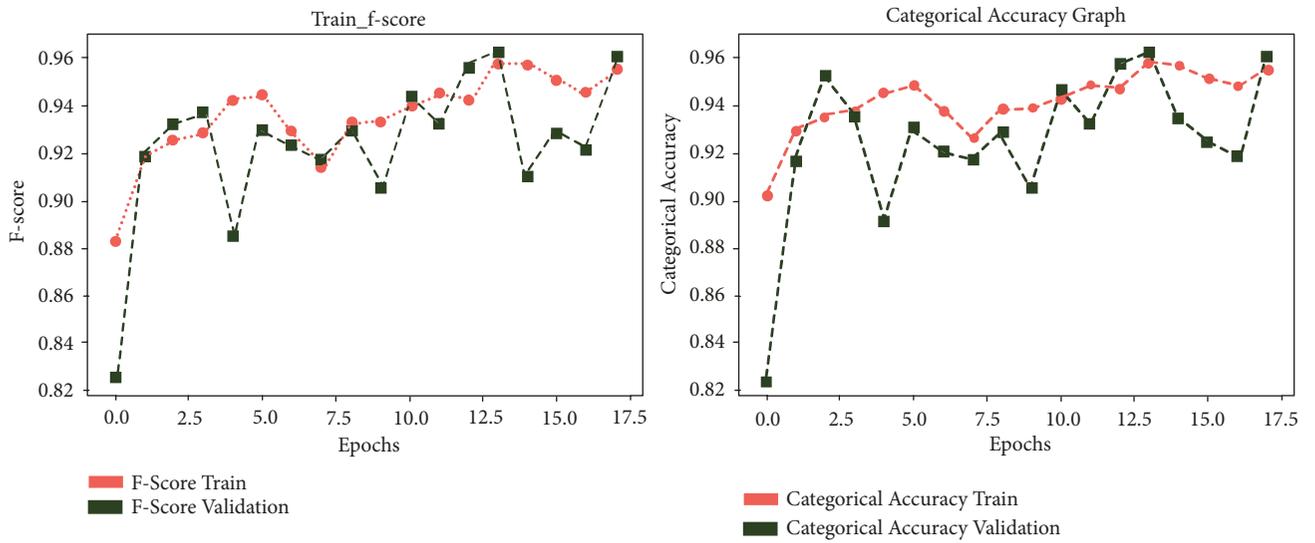

Figure 5: F-beta score and categorical accuracy GRU.

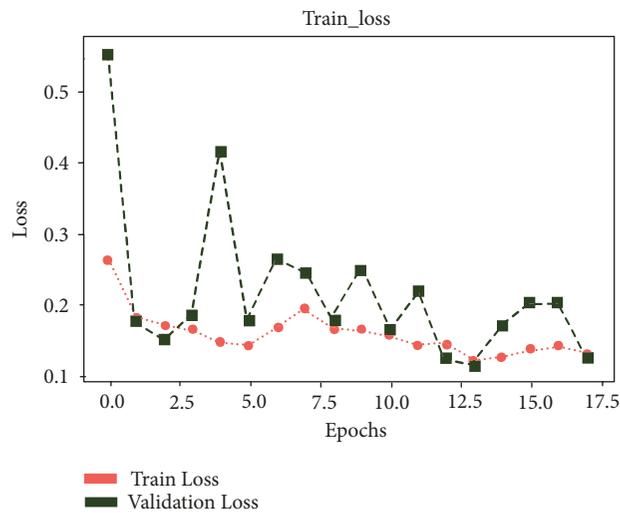

Figure 6: Categorical cross-entropy GRU.

similar to the one use for the creation of machine and deep learning models. Thus, the Python model will be deployed in a standard unit with a Port Mirroring from the router.

## Data Availability

The dataset used to support the findings of this study is available in https://joseaveleira.es/dataset. ©® reg#LE-229-18.

## Conflicts of Interest

The authors declare that they have no conflicts of interest.

## Acknowledgments

This work is partially supported by (i) Instituto Nacional de Ciberseguridad (INCIBE) and developed Research Institute of Applied Sciences in Cybersecurity (RIASC); (ii) Junta de Castilla y León, Consejería de Educación, Project LE078G18. UXXI2018/000149. U- 220.

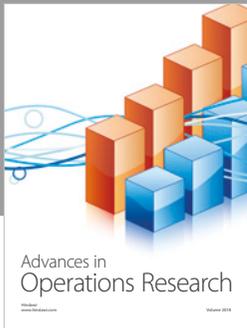
Advances in
**Operations Research**

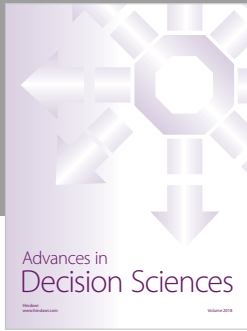
Advances in
**Decision Sciences**

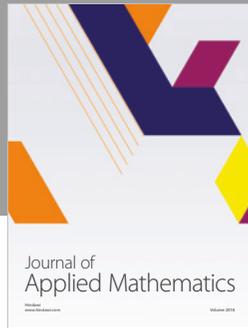
Journal of
**Applied Mathematics**

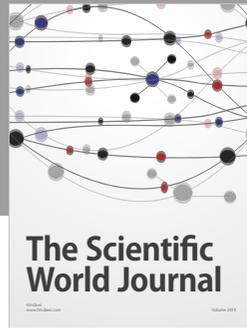
**The Scientific World Journal**

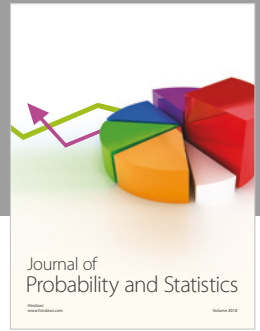
Journal of
**Probability and Statistics**

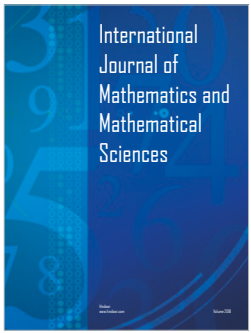
International Journal of
**Mathematics and Mathematical Sciences**

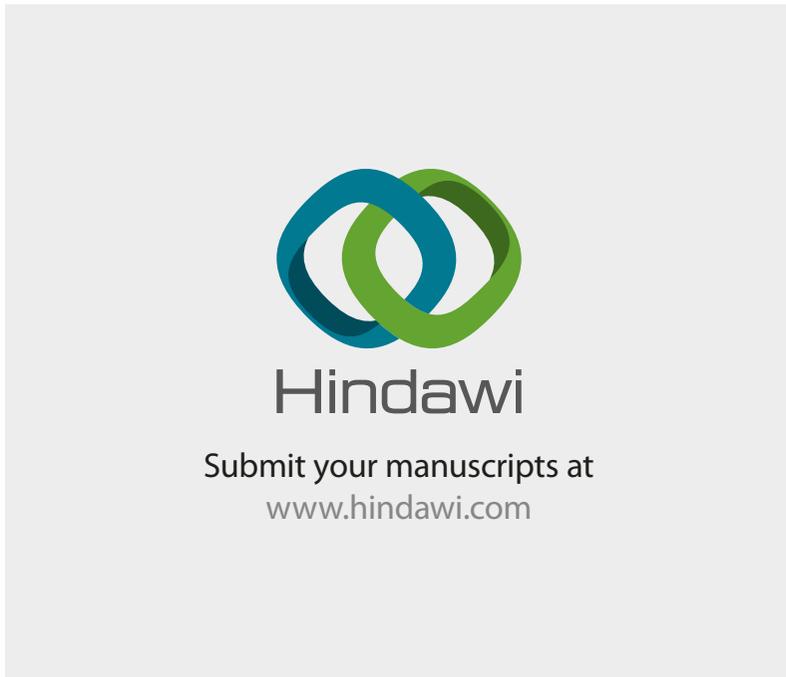
Submit your manuscripts at
www.hindawi.com

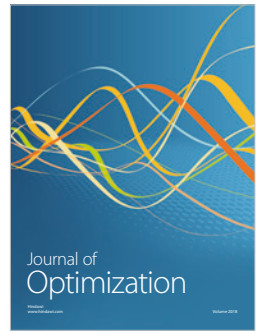
Journal of
**Optimization**

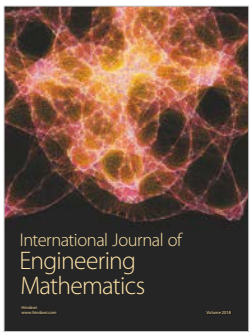
International Journal of
**Engineering Mathematics**

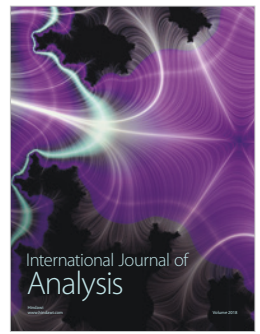
International Journal of
**Analysis**

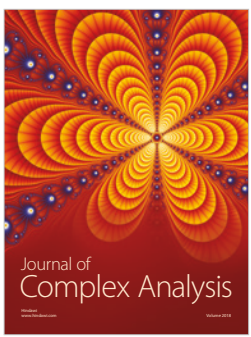
Journal of
**Complex Analysis**

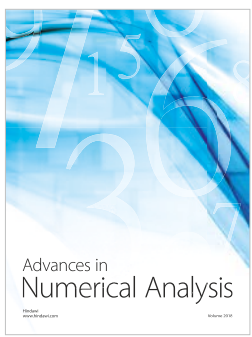
Advances in
**Numerical Analysis**

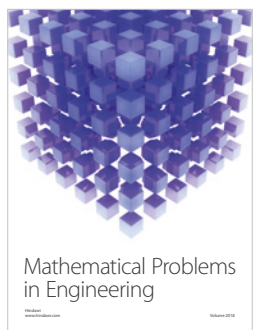
**Mathematical Problems in Engineering**

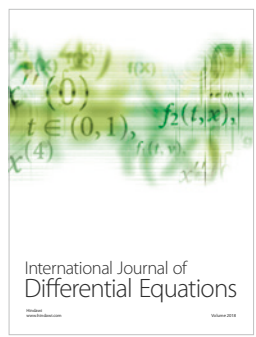
International Journal of
**Differential Equations**

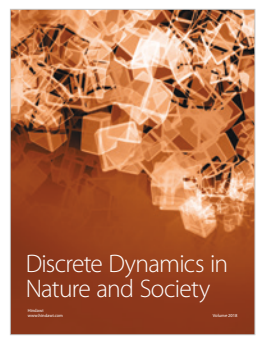
**Discrete Dynamics in Nature and Society**

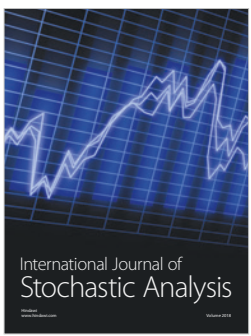
International Journal of
**Stochastic Analysis**

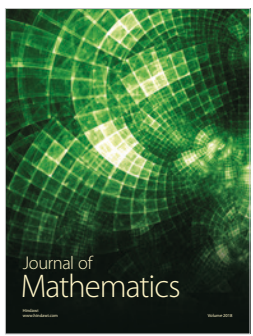
Journal of
**Mathematics**

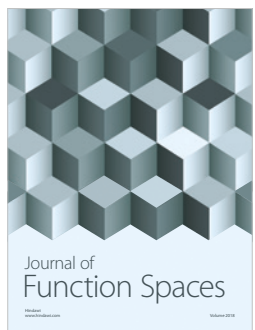
Journal of
**Function Spaces**

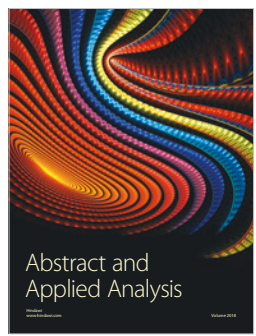
**Abstract and Applied Analysis**

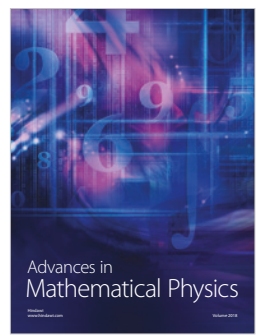
Advances in
**Mathematical Physics**